%% file: 0_Paper.tex
\documentclass[11pt,a4paper]{article}
\usepackage[hyperref]{emnlp2020}
\usepackage{times}
\usepackage{latexsym}
\usepackage{math-common}
\usepackage[shortlabels]{enumitem}
\usepackage{float}

\usepackage{microtype}
\usepackage{algorithm}
\usepackage{algpseudocode}
\usepackage{fancyvrb}
\usepackage{inconsolata}
\usepackage{graphicx}
\usepackage{arydshln}

\definecolor{ao}{rgb}{0.0, 0.5, 0.0}

\DefineVerbatimEnvironment{VerbOutC}{Verbatim}{fontfamily=zi4}

\aclfinalcopy 

\title{Autoregressive Knowledge Distillation through Imitation Learning}

\author{Alexander Lin$^1$\\
  ASAPP, Inc. \\
  New York, NY, USA \\
  \And
  Jeremy Wohlwend$^2$ \\
  ASAPP, Inc. \\
  New York, NY, USA \\
  \And
  Howard Chen$^3$ \\
 ASAPP, Inc. \\
  New York, NY, USA \\
  \And
  Tao Lei$^4$ \\
  ASAPP, Inc. \\
  New York, NY, USA \\
  \AND 
  $^1$\textnormal{\texttt{alexanderlin01@g.harvard.edu}} \\
  $^2$\textnormal{\texttt{jwohlwend@csail.mit.edu}}
  \And 
  $^3$\textnormal{\texttt{howardchen@cs.princeton.edu}} \\
  $^4$\textnormal{\texttt{tao@asapp.com}}
  }

\date{$^1$ \texttt{alexanderlin01@g.harvard.edu}}

\begin{document}
\maketitle

\begin{abstract}
The performance of autoregressive models on natural language generation tasks has dramatically improved due to the adoption of deep, self-attentive architectures.  However, these gains have come at the cost of hindering inference speed, making state-of-the-art models cumbersome to deploy in real-world, time-sensitive settings. 
We develop a compression technique for autoregressive models that is driven by an imitation learning perspective on knowledge distillation.  The algorithm is designed to address the exposure bias problem.    
On prototypical language generation tasks such as translation and summarization, our method consistently outperforms other distillation algorithms, such as sequence-level knowledge distillation.  Student models trained with our method attain 1.4 to 4.8 BLEU/ROUGE points higher than those trained from scratch, while increasing inference speed by up to 14 times in comparison to the teacher model.\footnote{Our code can be found at \url{https://github.com/asappresearch/imitkd}.}
\end{abstract}

\input{1_Introduction}
\input{2_Background}
\input{3_ImitKD}

\input{4_Related_Work}
\input{5_Experimental_Setup}

\input{6_Results}

\section{Conclusion}
In this work, we developed a new knowledge distillation technique inspired by imitation learning for compressing large and cumbersome autoregressive models into smaller and faster counterparts.  We demonstrated the empirical success of our method over popular baselines on several natural language generation tasks.

We are excited about several possible avenues for future work.  One branch of ideas involves incorporating more advanced IL algorithms beyond DAgger, such as LOLS \citep{chang2015learning}, to further improve the distillation process.  Another possibility is to design imitation-based fine-tuning analogs to the SeqInter method. 
Finally, although our experiments in this paper focused on sequence-to-sequence settings, we are interested in exploring the use of ImitKD for compressing large language models aimed at transfer learning.

\section*{Acknowledgments}
We thank the ASAPP NLP team -- especially Yi Yang, Nicholas Matthews, Joshua Shapiro, Hugh Perkins, Amit Ganatra, Lili Yu, Xinyuan Zhang, and Yoav Artzi --   as well as the EMNLP reviewers for their helpful feedback on the paper.  

\bibliography{anthology,emnlp2020}
\bibliographystyle{acl_natbib}

\clearpage
\newpage
\input{A_Appendix}

\end{document}

%% file: 1_Introduction.tex
\section{Introduction}
Autoregressive models are ubiquitous in natural language processing.  Due to the sequential nature of text generation, they are often the tool of choice for tackling sequence-to-sequence problems such as translation \citep{sutskever2014sequence}, summarization \citep{rush2015neural}, and dialogue \citep{eric2017copy}.  Furthermore, they form the backbone of several successful generative pre-training architectures \citep{howard2018universal, peters2018deep, radford2019better, dai2019transformer}.

Two recent trends have made autoregressive models cumbersome to deploy in real-world, natural language generation (NLG) applications.  First, state-of-the-art models have grown larger and larger, amounting to hundreds of millions and even billions of parameters \citep{dong2019unified, liu2019text, raffel2019exploring}.  The increase in size and depth dramatically slows down inference speed.  Second, the architecture of choice for autoregressive models seems to have shifted from the recurrent neural network (RNN) \citep{bahdanau2014neural, luong2015effective} to the Transformer \citep{vaswani2017attention}.  Though the Transformer's self-attention mechanism improves performance, it also increases the computational complexity of the step-by-step generation algorithms that are used at test time.  Thus, both of these trends have contributed to significantly increasing inference time costs, especially on CPUs and low-resource devices, hindering their use in production systems.


\emph{Knowledge distillation} (KD) \citep{bucilua2006model, hinton2015distilling} is one popular method for model compression.  It transfers the information learned by a large, pretrained \emph{teacher} to a smaller, untrained  \emph{student}.  In comparison to other methods such as weight pruning and quantization, KD allows the compressed model's architecture to significantly differ from that of the original teacher.  This feature enables models trained with KD to achieve high performance while meeting particular inference requirements (e.g. memory, speed, etc.).

\emph{Sequence-level knowledge distillation} (SeqKD), proposed by \citet{kim2016sequence}, is the dominant technique for autoregressive KD in the current NLG literature, especially for machine translation \citep{gu2017non, ren2019fastspeech, zhou2019understanding}.  This method trains a student model using a modified dataset generated by the teacher model and the standard negative log-likelihood objective.  While SeqKD is simple and efficient, we argue that it does not take advantage of the teacher's full potential.  



Training the student model with a \emph{static} dataset leads to the exposure bias problem. During training, the student model learns to predict the next token given previous tokens provided by the data. However, at inference time, the student generates the entire sequence from scratch by repeatedly using its own outputs as context for subsequent steps.  This training-inference inconsistency causes a decrease in generation quality.  Alternatively, we propose that the student can leverage the teacher in a \emph{dynamic} fashion during the learning process.



We devise a new compression algorithm for autoregressive models called \emph{imitation-based knowledge distillation} (ImitKD).  It is inspired by an imitation learning (IL) perspective on the autoregressive distillation problem.  Our algorithm trains a student model within an IL framework by treating the teacher as an oracle, and allows the student to explore its own generation during training.  The teacher corrects the student's generation at every time step, thereby guiding the student in learning how to generate.

Experimental results in translation and summarization show that ImitKD is especially suitable for compressing deep Transformer models that achieve high performance into shallow RNNs that generate up to 14 times faster at inference time.  Our method consistently outperforms other distillation algorithms (such as word-level KD and sequence-level KD), and yields student models that beat models trained without a teacher by 1.4 to 4.8 points on generation metrics such as BLEU and ROUGE.

%% file: 2_Background.tex
\section{Background}
\subsection{Autoregressive Distillation}
First, we formalize the task of autoregressive distillation.
An \emph{autoregressive model} $\pi$ specifies a joint distribution over a $T$-dimensional \emph{target sequence} $\bd y = \{y_1, \ldots, y_T\} \in \mathcal{Y}$ by decomposing it into a product of univariate conditionals:
\begin{align}
\pi(\bd y) = \prod_{t=1}^T  \pi(y_t \given \bd y_{<t}),
\end{align}
where $\bd y_{<t}$ denotes $\{y_1, \ldots, y_{t-1}\}$ for $t > 1$ and $\emptyset$ for $t = 1$.  The joint distribution over $\bd y$ may itself be conditional on some related \emph{source feature} $\bd x \in \mathcal{X}$ (e.g. translation, summarization) or not (e.g. language modeling).  
Since the former case can generalize the latter by letting $\mathcal{X} = \emptyset$, we will specify the presence of $\bd x$ in the rest of the paper. 

In autoregressive distillation, the goal is to learn a \emph{student model} $\pi$ that performs well at sequence generation by minimizing its loss with respect to a pre-trained \emph{teacher model} $\pi^*$.
In many cases, the training objective can be expressed as
\begin{align}
\Loss(\pi) =\, \E_{\,\bd y | \bd x \sim \mathcal{D}} \left[\sum_{t=1}^T \ell^{\pi^*}(\bd y_{<t}, \bd x; \pi) \right], \label{eq:dobj}
\end{align}
where $\ell^{\pi^*}(\cdot; \pi)$ is the next-token loss function measuring the discrepancy between the teacher and student models given some prior context $\{\bd y_{<t}, \bd x\}$.  

Here, $\mathcal{D}$ denotes a distribution (or dataset) of source-target pairs $\bd x \to \bd y$.  Due to the combinatorial nature of sequence generation, an autoregressive distillation method must maximize its learning efficiency by carefully $\mathcal{D}$, i.e. how it explores the exponentially-sized space.  We motivate this choice with the field of \emph{imitation learning}, an active research area of reinforcement learning.  


%

\subsection{Distillation as Imitation Learning}
Autoregressive text generation can be interpreted as a $T$-step Markov decision process (MDP).
In particular, the autoregressive model $\pi$ we wish to learn can be treated as a policy learner that maps a state to a distribution over actions.
In our case, a state is a partial sequence $\bd y_{<t}$ for $t < T$, an action is the next token $y_t$, and the action space is the vocabulary.  Given a state (partial sequence) and a chosen action (next token), the transition function is deterministic and simply concatenates them to form a new state (partial sequence). 

The policy learner must be trained using some form of supervision.
One option is to use reward-based reinforcement learning, which requires defining the numerical quality of a state.  However, for the autoregressive distillation problem, an arguably better choice is \emph{imitation learning} (IL), which optimizes the policy by learning from demonstrations.  In IL settings, an \emph{oracle} policy $\pi^*$ that is known to achieve high performance is provided during training.
As a result, we can recast the overall goal as minimizing the divergence of the policy $\pi$ from the oracle $\pi^*$. 
For example, it may be difficult to objectively define what it means for an aspiring translator to perform well at the local token-by-token level.  Yet, if we were given access to an expert translator, we could simply say the learner is performing well if they translate in the same way as the expert. 



The IL framework is well-suited for autoregressive distillation, since the student and teacher models naturally fill the respective roles of the learner $\pi$ and the oracle $\pi^*$.
Thus, we can easily apply theoretical results and practical methods from the IL literature to the autoregressive distillation problem.

\subsection{SeqKD as Behavioral Cloning} \label{tf-as-bc}
One distinguishing feature between different imitation learning methods pertains to how to define the state distribution $\mathcal{D}$ in the training objective (Equation~\ref{eq:dobj}).  Indeed, this is also one of the key design questions of autoregressive distillation.
For instance, one simple and effective IL method is \emph{behavioral cloning}~\cite{ross2010efficient}, which obtains $\mathcal{D}$ by running the oracle $\pi^*$ on the MDP.

The popular sequence-level knowledge distillation~(SeqKD) algorithm of~\citet{kim2016sequence} can be interpreted as behavioral cloning.
For each source feature $\bd x$ in the original training data, the teacher/oracle generates its (approximate) mode $\bd y^* = \arg\max_{\bd y'} \pi^*(\bd y' \given \bd x)$, typically using beam search.   This new set of $\bd x \to \bd y^*$ pairs forms a teacher-generated dataset $\mathcal{D}^*$ that serves as the state distribution for training the student.
In addition, the negative log-likelihood of the teacher's tokens $\bd y^* = \{y^*_1, \cdots, y^*_T\}$ is used as the loss $\ell^{\pi^*}(\cdot; \pi)$.  The overall training objective $\mathcal{L}_\text{SeqKD}(\pi)$ is
\begin{align}
\E_{\bd y^*|\bd x \sim \mathcal{D}^*} \left[\sum_{t=1}^T  -\log  \pi(y_t^*\given \bd y_{<t}^*, \bd x)\right]. \label{eq:seqkd}
\end{align}

The key advantage of SeqKD (as well as behavioral cloning) lies in its simplicity -- we only need some samples from the teacher/oracle to work with. 
In comparison to vanilla supervised learning (which minimizes the negative log-likelihood of human-generated text), SeqKD has no additional training overhead other than the creation of $\mathcal{D}^*$.

However, the simplicity of the algorithm also limits its potential.
\citet{ross2010efficient} argued that training a policy $\pi$ via behavioral cloning incurs regret with respect to the oracle $\pi^*$ that is a quadratic function of the time horizon $T$.
Intuitively, behavioral cloning suffers from the \emph{exposure bias problem}.  During training, the student model learns to perform good actions for the teacher/oracle's state distribution $\mathcal{D}^*$, but is never exposed to its own states.  Thus, during testing (when the student must walk an MDP of self-generated states), the step-by-step errors compound over time, resulting in suboptimal generations.

We argue that in autoregressive distillation, the teacher/oracle can do more than produce a static dataset.  It is a dynamic entity capable of interacting with the student throughout training.  By querying the teacher with its own states, the student has the opportunity to ameliorate exposure bias and learn how to generate. 



%% file: 3_ImitKD.tex
\section{Imitation-Based Distillation Algorithm}

In this section, we present our IL-based algorithm for autoregressive distillation.
We begin by describing the key design principles and why we expect them to work well.  Then, we elaborate on the algorithm's implementation in detail.

\subsection{Design Principles and Rationale}
One key principle of our algorithm is that the student model must be trained on its own state distribution so that it will perform better at generation.
In practice, we achieve this by sampling training examples from $\tilde{\mathcal{D}}$, a mixture of an initial distribution $\mathcal{D}$ (e.g. a static training set) and the distribution $\mathcal{D}_\pi$ of generations from the student $\pi$.  We use $\mathcal{D}$ to alleviate the cold-start problem, in which an untrained $\pi$ generates poorly at the start of training.  

This idea builds upon the empirical and theoretical foundation of dataset aggregation (DAgger), one of the most popular imitation learning methods that improve upon behavioral cloning.
DAgger~\cite{ross2011reduction} successively populates its training set by adding new data generated from the oracle-learner mixture.
It then re-trains the policy learner on the aggregated dataset at each iteration.
Under some assumptions (such as the loss function being strongly convex in $\pi$), \citet{ross2011reduction} proved that DAgger yields a policy $\pi$ that has linear regret in $T$ with respect to $\pi^*$.  
This is a significant improvement over the behavior cloning result and can be attributed to fixing exposure bias.
We expect a similar strategy of mixing oracle and learner distributions to work well for non-convex neural networks, as shown in other applications~\cite{zhang2016query,sun2017deeply}.

Another key principle of our algorithm is that the teacher model should play the role of the oracle and correct the student's generations at each time step.  In order for such a training strategy to be successful, the teacher must be able to provide better actions than the student for the student's own states.  To test this hypothesis, we experiment with a deep Transformer-based translation model completing the partial translations of a shallow RNN.
As shown in Table~\ref{tab:sanity}, the Transformer completions achieve much higher BLEU score than the RNN's full generations.
This validates our assumption that a strong teacher model can indeed play the role of the oracle and guide the student to better states.    

\begin{table}
\centering
\begin{tabular}{lc}
\hline
\textbf{Decoding Method} & \textbf{Bleu} $\uparrow$\\
\hline
Transformer only & 33.8 \\
\hline
RNN only & 28.6 \\
RNN first, Transformer completes& 31.4 \\
\hline
\end{tabular}
\caption{Greedy decoding BLEU scores on the IWSLT validation set for the preliminary test.}\label{tab:sanity}
\end{table}


\algblockdefx{MRepeat}{EndRepeat}{\textbf{repeat} }{}
\algnotext{EndRepeat}


\subsection{The ImitKD Algorithm}
Our imitation-based knowledge distillation algorithm (ImitKD) is given in Algorithm \ref{imit-alg}.  
The central training objective is
\begin{align}
\hspace{-0.5em} \Loss_{\text{ImitKD}}(\pi) = \E_{\bd y|\bd x \sim \tilde{\mathcal{D}}} \left[\sum_{t=1}^T \ell^{\pi^*}(\bd y_{<t}, \bd x; \pi) \right], \label{imitKD-loss}
\end{align}
where $\tilde{\mathcal{D}}$ is the data mixture defined by sampling from the initial dataset $\mathcal{D}$ and generating with the student (lines 8-11). 
The probability $\beta_i \in [0,1]$ (line 8) controls how often an example comes from $\mathcal{D}$.
The loss function $\ell^{\pi^*}$ can be realized as the negative log-likelihood of the oracle's optimal next token/action,
\begin{align}
\ell^{\pi^*}_{\text{opt}}(\bd y_{<t}, \bd x; \pi) = - \log \pi(v^* \given \bd y_{<t}, \bd x), 
\end{align}
where $v^* = \arg \max_{v \in \mathcal{V}} \pi^*(v \given \bd y_{<t}, \bd x)$.
Alternatively, $\ell^{\pi^*}$ can be the cross-entropy loss between the full distributions,
\begin{align}
&\ell^{\pi^*}_{\text{full}}(\bd y_{<t}, \bd x; \pi) \label{full} \\
&= -\sum_{v \in \mathcal{V}} \pi^*(v \given \bd y_{<t}, \bd x) \cdot \log \pi(v \given \bd y_{<t}, \bd x). \nonumber
\end{align}

\begin{algorithm}[!t!]
\caption{Imitation-Based Distillation} \label{imit-alg}
\begin{algorithmic}[1]
\State Let $\mathcal{D}$ be initial dataset.
\State Initialize $\pi_1$ at random.
\For {$i = 1, \ldots, I$}
    \State {Initialize new dataset $\tilde{\mathcal{D}}_i = \emptyset$.}
    \MRepeat {$B$ \textbf{times}}
        \State Sample an example $e = \bd y \given \bd x \sim \mathcal{D}$.
        \State Sample uniformly $u \sim [0, 1]$.
        \If {$u > \beta_i$} 
            \State Generate $\bhat{y}$ from $\pi_i$ given $\bd x$.
            \State Replace example with $e = \bhat y \given \bd x$.
        \EndIf
        \State Append example $e$ to $\tilde{\mathcal{D}}_i$.
    \EndRepeat
    \State Compute $\Loss_{\text{ImitKD}}(\pi_i)$ on $\tilde{\mathcal{D}}_i$ with $\pi^*$.
    \State Let $\pi_{i+1} = \pi_i - \alpha_i \cdot \partial \Loss_{\text{ImitKD}} / \partial \pi_i$.
\EndFor \\
\Return Best policy $\pi$ on validation set.
\end{algorithmic}
\end{algorithm}

Next, we describe some practical implementations in order to make Algorithm \ref{imit-alg} suitable for compressing deep learning systems.  
One limitation of DAgger is that the training data keeps growing, making each iteration successively more expensive.
As an alternative to aggregation, we perform \emph{data replacement} within each training batch.

As shown in Algorithm~\ref{imit-alg}, we treat each mini-batch $\tilde{\mathcal{D}}_i$ as a new iteration of the dataset
and perform a single step of stochastic gradient descent on $\Loss_\text{ImitKD}$ (Equation \ref{imitKD-loss}) with respect to the parameters of the previous model $\pi_i$ to yield $\pi_{i+1}$.
Thus, the number of iterations $I$ becomes the number of mini-batches used to train the student model.


Our practical algorithmic changes are inspired by theory.
The dataset aggregation algorithm \citep{ross2011reduction} achieves its regret bounds because it reduces to the Follow-the-Leader algorithm for online learning \citep{kakade2009complexity}.  Our training paradigm can be similarly interpreted as an online gradient descent algorithm, which has comparable guarantees for strongly convex losses \citep{hazan2007logarithmic} and even certain non-strongly convex losses \citep{garber2019logarithmic}.  Variants of this paradigm have also been employed in other deep learning work \citep{bengio2015scheduled, sun2017deeply}.

\subsection{Data Mixture Selection and Annealing}
Dataset replacement requires an initial dataset that can be potentially replaced at each step.  A natural candidate for this initial dataset is the original supervised training data (denoted as $\mathcal{D'})$, which can be interpreted as a collection of samples from a human oracle.
Alternatively, we can use the \mbox{SeqKD} dataset $\mathcal{D}^*$, which has generations from the teacher.

If we take samples from $\mathcal{D}'$ or $\mathcal{D}^*$ and replace some of them with student-generated samples, we effectively create a teacher-student dataset mixture.  Unlike DAgger, this mixture occurs at the sequence level instead of the token/state level.
An advantage of sequence-level mixtures is that they do not require generating with the teacher during each training iteration, which can be quite expensive if the teacher is a large neural network.  
Instead, the teacher only needs to compute the batched loss, which is comparatively much cheaper.  The exact mixing schedule $\beta_1, \ldots, \beta_I$ is a customizable feature of Algorithm \ref{imit-alg}.  Empirically, we have found an exponential decay to work well, i.e.
$\beta_i = r^{i / I}$,
where $r \in [0, 1]$ is the final mixing rate.

\subsection{Speeding Up Training}
Generating sequences $\bhat y$ on the fly at every iteration (line 9) can be a major computation bottleneck during training.
We speed up this step by generating a pool of $B \cdot M$ examples in parallel only once every $M$ iterations, where $B$ is the batch size and $M$ is a hyperparameter.
One caveat of this modification is that at iteration $i$, the loss function may no longer be computed on examples generated by the most recent set of model parameters, but rather parameters from up to $M$ iterations prior.  Nonetheless, we have found that setting $M$ to a small integer (e.g. 2-8) can speed up training time without impacting final model performance.

We use greedy decoding or top-$K$ sampling with small $K$ to produce samples $\bhat y$ (line 9) in our algorithm.
These two strategies are efficient to run, operate similarly to the generation employed at inference time, and have empirically worked well in our experiments.
Of course, the generation strategy can be customized for different tasks.



%% file: 4_Related_Work.tex
\section{Related Work}
\label{sec:length}

The distillation problem for autoregressive models was first tackled by \citet{kim2016sequence}, who introduced sequence-level knowledge distillation for neural machine translation.  Subsequent works have used seqKD for non-autoregressive translation models \cite{gu2017non, zhou2019understanding}, low-resource settings \citep{chen2017teacher}, and ensemble distillation with multiple teachers \citep{kuncoro2016distilling, tan2019multilingual}. \citet{wei2019imitation} proposed a behavioral cloning method for distilling autoregressive translation models into non-autoregresssive translation models.
In contrast, our method aims to address the learning challenges in autoregressive distillation, such as exposure bias.

Various methods other than standard supervised learning have been explored for training generative models of language.
MIXER~\cite{ranzato2015sequence} and Beam Search Optimization~\cite{wiseman2016sequence} also perform generation during training, but use sequence-level metrics (e.g. BLEU score) as training supervision.
Simlarly, SEARNN \cite{leblond2017searnn} trains RNNs to iteratively generate sequences with beam search to compute the local loss of a single action during the decoding process.
Scheduled sampling \citep{bengio2015scheduled} and its extensions \citep{goyal2017differentiable, zhang2019bridging} alleviate exposure bias by replacing some words in the true context with the model's prediction.
However, without a dynamic query-able oracle, these methods 
face the challenge of properly defining the training signal when the generated sequence no longer exists in the static training data.
For example, directly reusing the tokens in the static dataset as the target next token leads to an inconsistent training procedure~\cite{huszar2015not}.
In contrast to these methods, distillation can fully leverage the teacher oracle, allowing us to design a simple and efficient imitation learning algorithm.

%% file: 5_Experimental_Setup.tex

\section{Experimental Setup} \label{exp-settings}

We test our autoregressive distillation method and all baselines on three language generation tasks -- IWSLT 2014 German $\to$ English translation, WMT 2016 English $\to$ German translation, and CNN/DailyMail abstractive news summarization.


\paragraph{Datasets}
The IWSLT 2014 De$\to$En dataset consists of approximately 170K sequence pairs.  
Following standard practice~\citep{bahdanau2016actor, deng2018latent, wang2019improving},
we randomly sample 4\% of this dataset as the validation set and let the remaining be the training set.
The test set is the concatenation of the dev2010, tst2010, tst2011, and tst2012 files.
We use a shared vocabulary of 14K lowercased BPE tokens \citep{sennrich2015neural}.

The WMT 2016 En$\to$De dataset has 4.5 million training pairs.
We use the same preprocessing of the prior work~\citep{ott2018scaling}, newstest2013 as the validation set and newstest2014 as the test set.  
The vocabulary consists of 32K cased BPE tokens.

The CNN/DailyMail summarization dataset has 287K, 13K and 12K pairs in the training, validation and test sets, respectively.  Following prior work~\citep{see2017get}, we truncate documents to 400 tokens and summaries to 100 tokens in the training set.
During evaluation, we generate up to 128 tokens. 
We use a pre-trained BERT \citep{devlin2018bert} tokenizer with a vocabulary of 30K lowercased tokens \citep{liu2019text}.  

\paragraph{Models}  
Transformers often attain state-of-the-art performance on common language generation tasks.  
On the other hand, RNNs (without self-attention) generate much faster at inference time.  Thus, from a practitioner's standpoint, it may be most desirable to compress a high-performing Transformer into a lightweight RNN.  For all tasks, we use the state-of-the-art Transformer architecture~\cite{vaswani2017attention} as the teacher model.
The teacher models are trained using vanilla supervised learning.
For WMT, we directly use the pre-trained Transformer model provided by the Fairseq library~\cite{ott2018scaling, ott2019fairseq}.

In all tasks, we use a recurrent neural network, specifically SRU~\cite{lei2017simple}, as the student model.
For completeness, we also train Transformer, GRU \citep{cho2014learning}, and LSTM \citep{hochreiter1997long} based student models on the IWSLT translation task, illustrating the effectiveness of our distillation method for various neural architectures.  All RNN-based models follow the seq2seq, encoder-decoder architecture \citep{sutskever2014sequence} and employ a single scaled dot-product attention between the encoder and decoder~\cite{bahdanau2014neural, luong2015effective}.

All models are trained using the Adam optimizer~\cite{kingma2014adam} with an inverse-square-root learning rate scheduler and learning rate warmup~\cite{vaswani2017attention}.
Our experiments were conducted using Flambé, a PyTorch-based model training and evaluation library \citep{wohlwend2019flambe}.
More implementation details such as hyperparameter settings are provided in Appendix A.

\begin{table}
\centering
\begin{tabular}{llll}
\hline
\textbf{Variant} & \textbf{Context (States)} & \textbf{Loss}  \\
\hline
Vanilla & Data & NLL\\
SeqKD & Teacher & NLL\\
ImitKD & Student/Data Mix& NLL\\
ImitKD* & Student/Teacher Mix& NLL\\
\hdashline
Vanilla + Full & Data & $\ell_{\text{full}}^{\pi^*}$\\
SeqKD + Full & Teacher & $\ell_{\text{full}}^{\pi^*}$\\
ImitKD + Full & Student/Data Mix&  $\ell_{\text{full}}^{\pi^*}$\\
ImitKD* + Full & Student/Teacher Mix& $\ell_{\text{full}}^{\pi^*}$ \\
\hline
\end{tabular}
\caption{Summary of training variants.  Base variants use the negative log-likelihood (NLL) of the optimal next token -- which is taken from the data for Vanilla, found using beam search for SeqKD, and queried from the teacher for ImitKD (i.e. $\ell^{\pi^*}_\text{opt}$).  All ``+ Full" variants are trained with the full teacher-student cross entropy.}\label{tab:variants}
\end{table}

\paragraph{Variants}
For the student models, we compare a wide range of training variants, including baselines such as vanilla supervised learning (which directly uses the original training set) and sequence-level knowledge distillation (SeqKD).  All SeqKD variants form the teacher-generated dataset using beam search with beam size $K = 5$.
For our imitation-based method, we experiment with annealing from the original training set (ImitKD) or the teacher-generated \mbox{SeqKD} dataset (ImitKD$^*$).
We also experiment with different token-level losses; base variants are trained with the optimal next token while ``+ Full" variants are trained with the full cross entropy.  Table~\ref{tab:variants} summarizes all variants and highlights their differences.  Note that the Vanilla + Full baseline -- referred to as ``WordKD" by \citet{kim2016sequence} -- has appeared in other distillation works \citep[e.g.][]{tan2019multilingual, sanh2019distilbert}.



\paragraph{Evaluation}
We use BLEU score \citep{papineni2002bleu} for translation and report \mbox{ROUGE-1}, ROUGE-2 and ROUGE-L scores \citep{lin2004rouge} for summarization.  
For all models, the training checkpoint with the highest BLEU/ROUGE-1 score on the validation set is used for test set evaluation.
We also report the perplexity metric for all tasks.

%% file: 6_Results.tex
\section{Results}

\paragraph{IWSLT De$\rightarrow$En Translation}

Table~\ref{tab:iwslt} compares all distillation methods on the IWSLT dataset.
The teacher model is an 8-layer Transformer.  We use a 3-layer SRU, a 2-layer SRU and a 2-layer Transformer as student models.
For all three student models, our ImitKD method outperforms all baselines in terms of BLEU score with beam size 1 (Bleu$_1$), BLEU score with beam size 5 (Bleu$_5$) and perplexity (PPL).
The improvement on Bleu score ranges from 1.4 to 4.8 points compared to the Vanilla training method.
The 3-layer SRU model trained with ImitKD + Full even slightly exceeds the performance of the teacher model.  Furthermore, our method consistently outperforms SeqKD by up to 1.4 BLEU, highlighting the benefit of training the student model with its own state distribution.

To further demonstrate the effectiveness of \mbox{ImitKD} across different model types, we report validation set Bleu$_1$ for various 2-layer neural architectures in Table~\ref{tab:diff-models}.  Our ImitKD method outperforms the baselines in all cases, with the gains being especially large for recurrent architectures.

\begin{table}[h!]
\centering
\begin{tabular}{lrrr}
\hline
\textbf{Variant} & \textbf{PPL} $\downarrow$ & \textbf{Bleu}$_1$  $\uparrow$ & \textbf{Bleu}$_5$ $\uparrow$\\
\hline
\textbf{Transf. (8-layer)} \\
\quad Teacher & 5.6 & 34.4 & 35.2 \\
\hline
\textbf{SRU (3-layer)} \\
\quad Vanilla & 7.4 & 30.0 & 31.2 \\
\quad SeqKD & 153.0 & 33.0 & 33.1\\
\quad ImitKD & 14.7 & 34.1 & 34.4\\ \hdashline
\quad Vanilla + Full & 5.4 & 34.2 & 34.8\\
\quad SeqKD + Full & 6.1 & 34.3 & 34.8\\
\quad ImitKD + Full& \textbf{5.3} & \textbf{34.8} & \textbf{35.4}\\
\hline
\textbf{SRU (2-layer)} \\
\quad Vanilla & 7.4 & 29.5 & 30.6 \\
\quad SeqKD & 102.1 & 32.0 & 32.4\\
\quad ImitKD & 12.7 & 33.3 & 33.7\\ \hdashline
\quad Vanilla + Full & 6.0 & 33.0 & 33.8\\
\quad SeqKD + Full & 6.8 & 33.1 & 33.7\\
\quad ImitKD + Full& \textbf{5.7} & \textbf{33.7} & \textbf{34.5}\\
\hline
\textbf{Transf. (2-layer)} \\
\quad Vanilla & 6.4 & 32.8 & 33.4\\
\quad SeqKD & 23.4 & 34.0 & 34.2\\
\quad ImitKD & 7.5 & 34.3 & 34.6\\ \hdashline
\quad Vanilla + Full & \textbf{5.9} & 33.8 & 34.2\\
\quad SeqKD + Full & 7.1 & 34.0 & 34.4\\
\quad ImitKD + Full& \textbf{5.9} & \textbf{34.4} & \textbf{34.8}\\
\hline
\end{tabular}
\caption{Results on IWSLT test dataset.}\label{tab:iwslt}
\end{table}

\begin{table}[h!]
\centering
\begin{tabular}{lllll}
\hline
\textbf{Variant} & \textbf{SRU} & \textbf{GRU} & \textbf{LSTM} & \textbf{Transf.}  \\
\hline
Vanilla & 28.6 & 28.6  & 27.7 &  32.4 \\
SeqKD  &  31.4 & 31.2 & 30.5 & 33.3 \\
ImitKD &  \textbf{32.7} & \textbf{32.7} & \textbf{32.4} & \textbf{33.7} \\
\hline
\end{tabular}
\caption{BLEU scores of different student architectures on the IWSLT validation set. We use a beam size of $1$.  The teacher attains a validation BLEU of 33.8.}\label{tab:diff-models}
\end{table}

\paragraph{WMT En$\to$De Translation}
Table~\ref{tab:wmt} presents our results for the WMT dataset.
The teacher is a 6-layer Transformer and the student is a 4-layer SRU.  
Here, we see that ImitKD performs closer to SeqKD.
These results reveal that direct behavioral cloning (SeqKD) can be quite effective when the amount of oracle demonstrations is sufficiently high, e.g. several millions of examples.
Nonetheless, ImitKD and ImitKD* can improve on \mbox{SeqKD} by training the student with its own states.  Among all variants, ImitKD + Full performs the best while avoiding the overhead of creating a teacher-modified dataset.  Furthermore, we see that \mbox{ImitKD} is especially effective in low-data regimes.
As shown in the bottom block of Table~\ref{tab:wmt}, ImitKD methods achieve much stronger results over baselines when we reduce the WMT training data to the same size as IWSLT.

\paragraph{CNN/DailyMail Summarization}
In Table \ref{tab:cnn}, we present the CNN/DailyMail results for a 6-layer Transformer teacher and a 2-layer SRU student.
Once again, the best student is ImitKD + Full, which achieves ROUGE scores that are within 1 point of the teacher's.
ImitKD variants outperform the baselines on all ROUGE metrics, showcasing the utility of our method on a different NLG task.



\begin{table}[h!]
\centering
\begin{tabular}{lrrrr}
\hline
\textbf{Variant} & \textbf{PPL} $\downarrow$ & \textbf{Bleu$_{1}$} $\uparrow$ & \textbf{Bleu$_5$} $\uparrow$ \\
\hline
Teacher & 3.2 & 28.7 & 29.2 \\
\hline
Vanilla & 5.5 & 22.0  & 23.1 \\
SeqKD  &  9.0 & 24.9 & 25.5 \\
ImitKD &  7.4 & 24.6 & 25.5 \\
ImitKD* & 8.4 & \textbf{25.3} & 25.8  \\
\hdashline
Vanilla + Full & \textbf{5.2} & 23.8 & 24.5 \\
SeqKD + Full & 5.6 & 24.7 & 25.3\\
ImitKD + Full & 5.6 & \textbf{25.3} & \textbf{25.9}\\
ImitKD* + Full & 5.6 & 25.0 & 25.8\\
\hline
$^\triangle$Vanilla & 18.7 & 13.8  & 15.1  \\
$^\triangle$SeqKD & 42.3 & 17.1 & 17.9 \\
$^\triangle$ImitKD & \textbf{15.0} & 17.8 & 19.0 \\
$^\triangle$ImitKD* & 17.8 & \textbf{18.6} & \textbf{19.5}   \\
\hline
\end{tabular}
\caption{Results on WMT dataset.  ImitKD* is trained on a student/teacher dataset mixture.  $\triangle$ indicates that the model is trained with 25$\times$ less data.  } \label{tab:wmt}
\end{table}

\begin{table}[h!]
\centering
\begin{tabular}{lrrrr}
\hline
\textbf{Variant} & \textbf{PPL} $\downarrow$ & \textbf{R1} $\uparrow$ & \textbf{R2} $\uparrow$ & \textbf{RL} $\uparrow$ \\
\hline
Teacher & 12.5 & 39.0 & 17.6 & 35.7 \\
\hline
Vanilla & 14.7 & 36.1  & 15.6 & 32.8 \\
SeqKD  & 52.9 & 36.4 & 16.1 & 33.1 \\
ImitKD & 17.2 & 37.3 & 16.4 & 34.1 \\
ImitKD* & 37.1 & 37.7 & 16.7 & 34.5  \\ \hdashline
Vanilla + Full & \textbf{13.6} & 36.2 & 16.0 & 32.9 \\
SeqKD + Full & 20.2 & 37.4 & 16.5 & 34.0 \\
ImitKD + Full & 14.0 & \textbf{38.4} & \textbf{17.1} & \textbf{34.9} \\
ImitKD* + Full & 17.9 & 38.1 & \textbf{17.1} & 34.6   \\
\hline
\end{tabular}
\caption{Results on CNN/DailyMail dataset.  All models generate using beam search $K = 5$ decoding.} \label{tab:cnn}
\end{table}

\paragraph{Size and Speed Analysis}
In Table \ref{tab:performance}, we analyze how our distillation technique can reduce computational costs, using the IWSLT (Table \ref{tab:iwslt}), WMT (Table \ref{tab:wmt}), and CNN/DailyMail (Table \ref{tab:cnn}) teacher/student pairs as case studies.  By training small student models with ImitKD, we can substantially decrease model size and increase inference speed, while minimizing performance loss.  
Shallow, recurrent architectures are especially attractive, because they can generate 4-14 times faster than deep Transformer teachers, and 2-3 times faster than Transformer students of similar size.

\begin{table*}[h!]
\centering
\begin{tabular}{llrlrlr}
\hline
\textbf{Task} & \textbf{Model} & \textbf{Size} & \textbf{\% Compress} \hspace{-2em} & \textbf{CPU Time} & \textbf{$\times$ Faster} & \textbf{\% Perform}\\
\hline
IWSLT & Transf. (8-layer)$^\dagger$ & 20.1 M & | & 269 / 518 ms & | & |\\
& SRU (3-layer) & 14.0 M & 68.7\% & 55 / 97 ms & 4.9 / 5.3 & 101.2\% \\
& SRU (2-layer) & 8.6 M & 42.7\% & 37 / 56 ms & 7.2 / 9.3 & 98.0\%\\
& Transf. (2-layer) & 8.5 M & 42.3\% & 78 / 144 ms & 3.4 / 3.6 & 100.0\%\\
\hline
WMT & Transf. (6-layer)$^\dagger$ & 209.9 M & | & 816 / 1466 ms & | & |\\
& SRU (4-layer) & 34.2 M & 16.3\% & 174 / 306 ms & 4.7 / 4.8 & 88.2\%\\
\hline
CNN/DM & Transf. (6-layer)$^\dagger$ & 59.8 M & | & 1900 / 12138 ms & | & |\\
& SRU (2-layer) & 14.4 M & 24.1\% & 258 / 826 ms & 7.4 / 14.7 & 98.5\%\\
\hline
\end{tabular}
\caption{Model types, along with statistics on size (in number of parameters) and CPU inference time per decoding (in milliseconds) for beam search with beam size $K \in \{1, 5\}$.  Teacher models are marked with $^\dagger$.  The ``\% Perform" column records the ratio of the best student's performance to the teacher's performance on BLEU score for translation tasks and ROUGE-1 for the summarization task.}\label{tab:performance}
\end{table*}

\begin{figure*}[h!]
    \centering
    \includegraphics[scale=0.42]{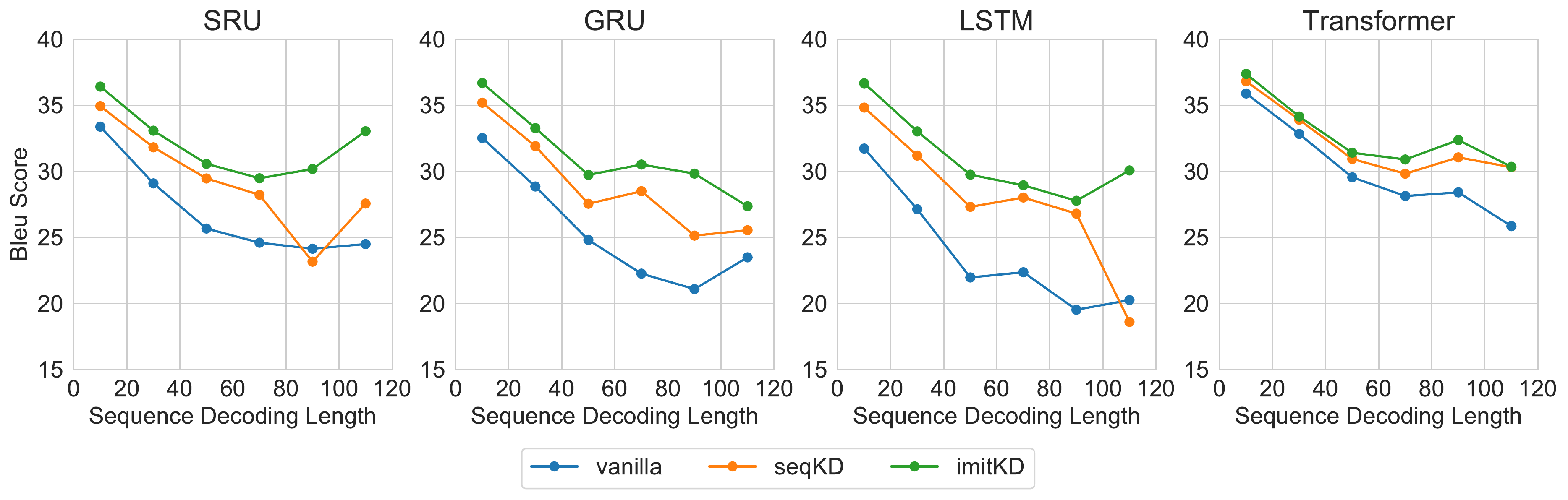}
    \caption{Bleu score versus sequence decoding length across different models and training variants.  Each point on the graph represents the Bleu score of all sequences whose length is within a bin of width 20.}
    \label{fig:length}
\end{figure*}

\paragraph{Performance Analysis at Different Lengths}
Figure \ref{fig:length} breaks down BLEU score vs. decoding length for IWSLT models trained with different algorithms (Vanilla, SeqKD, ImitKD).
We show results for the three types of RNNs and the Transformer of Table~\ref{tab:diff-models}.  All models have two layers.

As expected, we observe that the generation quality (in terms of BLEU score) degrades as the decoding length increases. 
This phenomenon can be explained by the global error compounding with each additional decision step \cite{ross2011reduction} and has been reported in previous works~\citep{bahdanau2014neural,zhang2019bridging}.  
As shown in Figure \ref{fig:length}, models trained with the vanilla objective, especially RNN-based models, suffer the most from this problem.
SeqKD improves the performance across all sequence lengths, but still experiences some BLEU score degradation for longer sequences.  
\mbox{ImitKD} further improves the BLEU score across all bins, and more importantly, the improvement is most significant for longer sequences.  
This analysis suggests that ImitKD explicitly addresses the exposure bias problem for training student models.


\begin{table}
\centering
\begin{tabular}{lllll}
\hline
\textbf{Variant} & \textbf{SRU} & \textbf{GRU} & \textbf{LSTM} & \textbf{Transf.}  \\
\hline
Vanilla & 32.1 & 31.9  & 31.2 &  34.0 \\
SeqKD  &  32.2 & 32.0 & 31.4 & 34.2 \\
ImitKD &  \textbf{33.5} & \textbf{33.4} & \textbf{33.1} & \textbf{34.4} \\
\hline
\end{tabular}
\caption{IWSLT validation set BLEU scores of \mbox{SeqInter} fine-tuning applied to the different student architectures of Table \ref{tab:diff-models}.  We use a beam size of 1.  The teacher (without fine-tuning) attains a validation BLEU of 33.8.}\label{tab:fine-tuning}
\end{table}    

\paragraph{Additive Effect of Fine-Tuning} \citet{kim2016sequence} propose a fine-tuning method for autoregressive distillation called SeqInter.  This method can further improve pretrained student models by exposing them to the sequence in the teacher beam's that is closest to the target in terms of sentence-level BLEU.  In Table \ref{tab:fine-tuning}, we show the results of applying SeqInter to each of the IWSLT models that were trained from scratch in Table \ref{tab:diff-models}.  While \mbox{SeqInter} enables Vanilla models to ``close the gap" on \mbox{SeqKD} models, ImitKD models clearly maintain their superior performance even after fine-tuning.



%% file: A_Appendix.tex
\appendix

\section{Appendices}
\label{sec:appendix}

\subsection{DAgger Algorithm}
The dataset aggregation (DAgger) algorithm \citep{ross2011reduction} minimizes the following objective:
\begin{align}
\mathcal{L}_{\text{Imit}}(\pi) = \E_{s_1, \ldots, s_T \sim \mathcal{D}} \left[\sum_{t=1}^T \ell^{\pi^*}(s_t; \pi)\right],
\end{align}
where $\mathcal{D}$ is a distribution (or dataset) of $T$-step state trajectories and $\ell^{\pi^*}(s, \pi)$ is the action-discrepancy loss between the oracle $\pi^*$ and the policy learner $\pi$ in state $s$.  The full DAgger algorithm is given in Algorithm \ref{dagger}.

\begin{algorithm}[H]
\caption{Dataset Aggregation} \label{dagger}
\begin{algorithmic}[1]
\State Let $\mathcal{D} = \emptyset$ be initial dataset.
\State Initialize $\pi_1$ at random.
\For {$i = 1, \ldots, I$}
    \State Let mixture policy $\tilde{\pi}_i = \beta_i \pi^* + (1 - \beta_i) \pi_i$.
    \State {Initialize new dataset $\mathcal{D}^i = \emptyset$.}
    \MRepeat {$B$ \textbf{times}}
        \State Run MDP on $\tilde{\pi}_i$, sample $\{s_1, \ldots, s_T\}$.
        \State Append new states $\{s_1, \ldots, s_T\}$ to $\mathcal{D}^i$.
    \EndRepeat
    \State Aggregate $\mathcal{D} = \mathcal{D} \cup \mathcal{D}^i$.
    \State Train $\pi_{i+1}$ on $\mathcal{D}$ to min $\Loss_{\text{Imit}}$ with $\pi^*$.
\EndFor \\
\Return Best policy $\pi$ on validation set.
\end{algorithmic}
\end{algorithm}

\subsection{Implementation Details}
In all experiments, all RNN-based models with hidden dimension $N$ consist of a bidirectional encoder with hidden dimension $N / 2$ and a left-to-right decoder with hidden dimension $N$.

For BLEU score evaluation, we use the NLTK library.\footnote{\url{https://www.nltk.org/_modules/nltk/translate/bleu_score.html}}  For ROUGE score evaluation, we use the py-rouge library.\footnote{\url{https://github.com/Diego999/py-rouge}}

\paragraph{Preliminary Study}
For Table \ref{tab:sanity}, we train both an 8-layer Transformer and a 2-layer RNN (specifically SRU) on the IWSLT dataset using standard supervised learning.  The architectural and training details are the same as those outlined in the IWSLT experiments.  At test time, both the Transformer and the RNN perform greedy decoding.  On average, ground-truth translations in the IWSLT test set have 24.5 tokens.  The ``RNN first, Transformer completes" mixed decoding strategy generates 12 tokens (i.e. half on average) with the RNN and the rest with the Transformer.  We measure generation quality using Bleu score.  

\paragraph{IWSLT}
The IWSLT 2014 German $\to$ English dataset is taken directly from the source website.\footnote{\url{https://sites.google.com/site/iwsltevaluation2014/data-provided}}

We train an 8-layer Transformer teacher model with model dimension 256, feedforward dimension 1024, and four attention heads as the teacher model.  
The 2-layer student SRU model has a hidden dimension 512, and the 3-layer model has hidden dimension 1024 and projection dimension 256.
The student Transformer model has model dimension 256, feedforward dimention 768 and 4 attention heads.

All models have word embedding dimension 256 and exhibit weight tying between the decoder embeddings and the output layer \citep{press2016using}.  
We train models for 80K steps with batch size 128 using the Adam optimizer with base learning rate 0.1.  
We use an inverse-square-root learning rate scheduler \citep{vaswani2017attention} with 10K warmup steps for the teacher and 5K warmup steps for all students.  Validation set metrics are recorded every 1K steps.
For all ImitKD variants, we set the final mixing rate $r = 0.005$ (i.e. very close to 0), and use top-$K$ sampling with $K = 5$ as the generation algorithm during training.  We use $M = 4$ as the batch parallelization parameter.  

In Table \ref{tab:diff-models}, the 2-layer SRU and the 2-layer Transformer follow the same architecture as those in Table \ref{tab:iwslt}.  To standardize architecture across RNNs, the GRU and the LSTM have the same embedding dimension (i.e. 256) and hidden dimension (i.e. 512) as the SRU.   

\paragraph{WMT}
The WMT 2016 dataset is taken from the Fairseq library.\footnote{\url{https://github.com/pytorch/fairseq/tree/master/examples/translation}}

We use a pre-trained Transformer-large model from the Fairseq library \citep{ott2018scaling, ott2019fairseq} as our teacher model.  It has embedding dimension 1024, model dimension 1024, and feedforward dimension 4096.  The student is a 4-layer SRU with hidden size 1024, projection size 256, and embedding size 256.  The student is trained for 15 epochs with batch size 512, base learning rate 0.1, and 4K warmup steps.  We record validation metrics every 1/4 of the epoch.  The encoder embeddings, decoder embeddings, and decoder output layer share the same weight parameters.
We tune the final mixing rate $r \in \{0.5, 0.1, 0.005\}$ for our ImitKD variants.

\paragraph{CNN/Dailymail}
The CNN/DailyMail dataset is taken from Professor Kyunghyun Cho's website, a commonly used source for this dataset.\footnote{\url{https://cs.nyu.edu/~kcho/DMQA/}}  
The teacher model is a 6-layer Transformer-base model with embedding dimension 512, model dimension 512, and feedforward dimension 2048.  
The student is a 2-layer SRU with embedding dimension 256, hidden size 1024, and projection size 256. 
We use a batch size of 128.  For both models, the learning rate follows an inverse-square root schedule with warmup of 2K steps.  
Validation set metrics are recorded every 2K steps.  
The teacher has a base learning rate of 0.03, while the student has a base learning rate of 0.1.  
The teacher benefits from larger effective batch sizes by accumulating gradients every eight steps.  
On the other hand, the student does not seem to benefit from gradient accumulation and therefore takes a gradient step after processing each batch.
All \mbox{ImitKD} variants use final mixing rate $r = 0.1$ and greedy decoding during training.  We use $M = 4$ as the batch parallelization parameter.    

\paragraph{Size and Speed Analysis}
CPU generation times for all models were measured on a 2019 MacBook Pro with a 2.6GHz 6-core Intel Core i7 processor.  Time estimates reported in Table \ref{tab:performance} were averaged over examples in the test set of the corresponding dataset. 

\paragraph{Performance Analysis at Different Lengths}
For each IWSLT variant, we ran greedy decoding (i.e. beam search decoding with beam size $K = 1$) on the test set.  Then, we sorted the decoded sequences by length into the following bins: 
[0, 20], [21, 40], [41, 60], [61, 80], [81, 100], [101, 120].  Each point in Figure \ref{fig:length} is the Bleu score of all sequences within one of these bins for the corresponding IWSLT variant.  

\paragraph{Additive Effect of Fine-Tuning}
For the fine-tuning experiments, we generated SeqInter data with a beam size of $K = 5$ and NLTK's sentence-level BLEU implementation.  We used the Adam optimizer with a base learning rate of 0.01 and an inverse-square root scheduler with 2K warmup steps.  All models were fine-tuned for 20K iterations.  Models were validated every 1K iterations.

%
%